\documentclass[preprint,12pt,authoryear]{elsarticle}

\usepackage{amssymb}
\usepackage{amsmath}

\usepackage{amsmath,amssymb,amsfonts}
\usepackage{algorithmic}
\usepackage{graphicx}
\usepackage{textcomp}
\usepackage{xcolor}
\usepackage{booktabs}
\usepackage[ruled,vlined]{algorithm2e}
\usepackage{multirow}
\usepackage{soul}
\usepackage{bbm}
\usepackage{amsthm}
\usepackage{subcaption}
\usepackage{tcolorbox}

\usepackage{natbib}
\setcitestyle{numbers,square}

\journal{Information Fusion}

\begin{document}

\begin{frontmatter}



\title{SparseDoctor: Towards Efficient Chat Doctor with Mixture of Experts Enhanced Large Language Models} 



\author{Jianbin Zhang$^{a}$, Yulin Zhu$^{a}$, Wai Lun Lo$^{a}$\footnote{Prof. Wai Lun Lo is the corresponding author.}, Richard Tai-Chiu Hsung$^{a}$, Harris Sik-Ho Tsang$^{a}$, and Kai Zhou$^{b}$} 

\affiliation[1]{organization={Department of Computer Science, Hong Kong Chu Hai College},
            city={Hong Kong},
            country={China}}

\affiliation[2]{organization={Department of Computing, The Hong Kong Polytechnic University},
            city={Hong Kong},
            country={China}}



\begin{abstract}

Large language models (LLMs) have achieved great success in medical question answering and clinical decision-making, promoting the efficiency and popularization of the personalized virtual doctor in society. However, the traditional fine-tuning strategies on LLM require the updates of billions of parameters, substantially increasing the training cost, including the training time and utility cost. To enhance the efficiency and effectiveness of the current medical LLMs and explore the boundary of the representation capability of the LLMs on the medical domain, apart from the traditional fine-tuning strategies from the data perspective (i.e., supervised fine-tuning or reinforcement learning from human feedback), we instead craft a novel sparse medical LLM named \textit{SparseDoctor} armed with contrastive learning enhanced LoRA-MoE (low rank adaptation-mixture of experts) architecture. To this end, the crafted automatic routing mechanism can scientifically allocate the computational resources among different LoRA experts supervised by the contrastive learning. Additionally, we also introduce a novel expert memory queue mechanism to further boost the efficiency of the overall framework and prevent the memory overflow during training. We conduct comprehensive evaluations on three typical medical benchmarks: CMB, CMExam, and CMMLU-Med. Experimental results demonstrate that the proposed LLM can consistently outperform the strong baselines such as the HuatuoGPT series. 

\end{abstract}




\end{frontmatter}



\section{Introduction}
\label{sec-intro}
Due to the rapid evolution of the \underline{L}arge \underline{L}anguage \underline{M}odels (LLM) in the natural language processing field, a series of universal chatbots have been developed in the real world to serve as the personalized secretary for human beings. Those chatbots possess powerful comprehension and analysis capabilities, as well as can naturally communicate with human beings in a human-friendly manner. For example, OpenAI crafted an epoch-making large language model named ChatGPT, which revealed a revolution in artificial intelligence. Based on the self-attention mechanism of the transformer layer, ChatGPT can capture the long-range dependencies between different tokens in the context of a corpus and thus can understand a large amount of text data at a high-level. The existence of ChatGPT and other GPT series has significantly influenced human society and has the potential to revolutionize other fields, including finance, education, marketing, medicine, and etc. 

After the giant success of the large language models, scientists endeavor to uncover the inherent causality under their strong understanding and inference capability from a data-driven perspective. It has been widely investigated that the strong natural language comprehension capability mainly comes from the two phenomena, i.e., emergence and homogenization. Emergence illustrates that the powerful understanding capability of LLMs on text data comes from the scaling up of the model size (number of parameters), training data and etc. For example, GPT-3 has more than 1750 billion parameters and thus presents incredibly powerful learning and reasoning capability for the text data. On the other hand, homogenization describes the large language models' strong adaptability across different domains and downstream tasks. Under these circumstances, some artificial general intelligence, such as ChatGPT, can accomplish some domain-specific tasks, like question answering in the medical field and communicate with users like an experienced doctor. 

However, there always exists a performance gap when we endeavor to use artificial general intelligence to solve domain-specific problems. For instance, ChatGPT sometimes will produce ``strange" and even wrong answers to misguide the patient and may cause a humanitarian catastrophe. We often call this phenomenon ``hallucination" of the LLM. It has been widely explored that the core reason for the hallucination problem is due to the lack of sufficient clinic-related text data during the training of the universal LLMs. Therefore, a series of medical LLMs~\cite{huatuogpt, HuatuoGPTII, doctorglm, chatdoctor, surveyMedicalLLM} has been established to bridge this gap and achieve significant performance gain on the medical question answering task. Specially, those medical LLMs highly rely on building an affluent medical-tailored corpus to amplify the influence of the clinic-related terminology, phrases, and logic during the fine-tuning phase of the LLM. For example, HuatuoGPT~\cite{huatuogpt} utilized the supervised fine-tuning strategy through synergizing the distilled data from the universal LLMs and collecting data from the doctors to prevent the ``model collapse"~\cite{mode_collapse} issue. Furthermore, HuatuoGPT-II~\cite{HuatuoGPTII} integrates the two-stage training process, including the continued pre-training and supervised fine-tuning, into a one-stage domain adaptation protocol to tackle the heterogeneous data from two distinct resources. Obviously, the researchers chose to utilize the data augmentation techniques and inject extra heterogeneous clinic-related data to force the LLMs to focus more on the clinical corpus during the supervised fine-tuning.  

However, the current medical LLMs solely endeavor to craft a medical LLM from a data-driven perspective, and omit the potential imperfections from the LLM's architecture, such as the bottleneck due to the high computational overhead during inference. Naturally, an interesting question comes up: \textit{``How to efficiently distill the clinical knowledge into a medical LLM from an architecture-driven perspective?"} 
Specifically, a series of promising works have been contributed to boost the efficiency of increasing the LLM's model size, i.e., drastically increasing the model size with a limited computational overhead increment. Parameter-efficient fine-tuning (PEFT) has emerged as a prominent paradigm in recent research. For instance, LoRA (low-rank adapter)~\cite{lora} utilizes a low-rank decomposition technique to split the original parameters of the transformer layer~\cite{transformer} into two trainable low-rank matrices. LoRA+~\cite{lora+} further introduces distinct learning rates for the updates of the two trainable low-rank matrices to improve the feature learning. Next, DoRA~\cite{DoRA} decomposes the pre-trained weight into two components, magnitude and direction, for fine-tuning to avoid any additional inference overhead. On the other hand, a line of works resort to the mixture of experts (MoEs)~\cite{MoE, gshard, switch_transformer, base_layers} technical route to achieve efficient fine-tuning of LLMs. Moreover, some researchers step further to combine the LoRA with MoE by introducing multiple LoRA experts with a router to select appropriate experts while freezing the large model~\cite{MOELoRA, mixlora, lorahub, loramoe, MoCLE}. Although the current LoRA-MoE experts theoretically combine the efficiency of the low rank decomposition with the capacity expansion of MoE, they still encounter three
major challenges: \textbf{C1)} A single adapter struggles to capture the inherent distinctions between the inter-task scenarios and leads to interference in the representation learning among different tasks. \textbf{C2)} The router exhibits weak preferences for the determination of activating the appropriate experts and produces indistinguishable representations across experts. \textbf{C3)} Traditional load-balancing strategies encourage uniform expert usage. However, the excessive balancing scheduling policy will reduce the routing confidence and harm the rationality of expert selection. 

Thereafter, these limitations inspire the breakthrough of our proposed fine-grained load-balancing enhanced LoRA-MoE. It is worth noting that we are the first to introduce the PEFT- and MoE-related techniques into the medical LLM framework and expand the current capability boundaries of medical LLM on clinical-related question answering tasks. Specially, our proposed medical LLM framework contains multiple LoRA experts, a sparse router, and a contrastive learning module on top of a large language model architecture to meticulously control the load balancing between each expert. To address the first challenge (\textbf{C1}), we introduce the MoE architecture on top of the open-source large language model and deploy the low-rank decomposition on each of the expert to strengthen the representation capability of the medical LLM on different tasks. For the second and third challenge (\textbf{C2} and \textbf{C3}), we introduce a novel contrastive learning framework supervised by a crafted expert contrastive loss to improve the routing mechanism of the current MoE. In the contrastive learning framework, we generate a \textit{routed expert view} together with a \textit{fused expert view}, which serves as the data augmentation for sake of forming the feature alignment between positive samples. Unlike the traditional expert contrastive learning method~\cite{MOELoRA} regards the outputs of the same expert as positive samples and the outputs of different experts as negative ones, ours instead regards the output features from the same token as the positive samples and vice versa. Under these circumstances, we could significantly tell apart different tokens of the clinical corpus in the latent space and improve the representation capability of the MoE system. Moreover, we also craft an expert memory queue mechanism to address the memory overflow issues in large-scale contrastive learning of LLM. In summary, our contributions are three-folds:

\begin{itemize}
    \item To the best of our knowledge, we are the first to investigate the enhancement of the question answering performance of LLM on the medical domain from the architecture perspective, instead of the data perspective.
    \item We introduce the novel contrastive learning enhanced low-rank MoE architecture to significantly improve the efficiency and performance of medical LLM on the clinical-related question answering downstream tasks.
    \item We conducted extensive experiments on different medical benchmarks, which demonstrate that ours can consistently outperform other strong baselines based on different medical benchmarks.  
\end{itemize}

\section{Related Works}
\subsection{Medical LLM}
For the medical domain, researchers have developed a variety of specialized LLMs fine-tuned from the universal LLMs, such as HuatuoGPT series~\cite{huatuogpt, HuatuoGPTII}, DISC-MedLLM~\cite{DISC-MedLLM}, and ChatDoctor~\cite{chatdoctor}, which achieve outstanding next token generation performance on English datasets (MedQA~\cite{LLM-MedQA}, MedMCQA~\cite{MedMCQA}) and Chinese datasets (CMB~\cite{CMB}, CMExam~\cite{CMExam}, CMMU\_Med~\cite{CMMU}). In particular, HuatuoGPT-II~\cite{HuatuoGPTII} integrates the continued pre-training framework and the supervised fine-tuning strategy into a single-stage training protocol, which avoids the dual distribution shifts and catastrophic forgetting issues of the traditional two-stage pipeline. It attains state-of-the-art performance on multiple Chinese medical evaluations. Nevertheless, such a model highly relies on full-parameter tuning or large-scale continued pre-training, which leads to high training cost and limited adaptation flexibility.

\subsection{MoE in LLM}
Mixture-of-Experts (MoE), first proposed by Jacobs et al. in 1991~\cite{6797059}, is a supervised learning framework that divides tasks among multiple expert networks and combines them via a gating network. In modern neural networks, MoE typically configures multiple experts in the
transformer layer~\cite{transformer} and adopts sparse gated routing that endeavors to activate only a few experts to participate in computation, thereby expanding model capacity without proportionally increasing the computational cost. For instance, the sparsely gated MoE proposed by Shazeer et al.~\cite{MoE} utilized the top-$k$ routing to maintain sparsity in both the training and inference phases, and combined it with a crafted load balancing loss to ensure even usage among different experts. This approach enabled RNN-based models such as \cite{transformer, switch_transformer, base_layers} to reach 137B parameters while keeping actual FLOPs at a low level because
only a few experts are involved in computation.
As the model scale expands, prior works such as GShard~\cite{gshard}, switch transformer~\cite{switch_transformer}, BASELayer~\cite{base_layers} further explore more advanced training strategies and routing mechanisms to integrate MoE with a series of large-scale transformers. These studies indicate that with a properly designed router and load balancing mechanism, MoE can significantly improve the representation capability of the language modeling and machine translation performance under limited computational resources. However, the current MoE frameworks also encounter the load imbalance and random routing problem, i.e., the router may tend to assign a large number of tokens to a few ``hot" experts and cause a significant waste of resources. Moreover, even under the balanced loads, the router may randomly select experts without preference, which leads to near-identical representations across experts. Thus, these bottlenecks will drastically hinder the diversity and specialization among the experts of MoE and lead to suboptimal performance on various downstream tasks, such as question answering, etc.

\section{preliminaries}
\subsection{Medical Question Answering}
Medical question answering (MedQA) leverages the decoder-only large language models to generate accurate, context-aware answers to clinical queries. The process can be mathematically formulated as a sequence-to-sequence (Seq2Seq) task with probabilistic constraints. Suppose the clinical question is tokenized into a sequence $\mathbf{X}=\{x_1,...,x_n\}$, which will be fed into a decoder-only LLM, i.e, $\mathbf{F}(\cdot)$, in order to generate an answer sequence $\mathbf{Y}=\{y_1,...y_m\}$. Intuitively, it can be formulated as the next token generation problem:
\begin{equation}
    P(y_t|\hat{\mathbf{Y}}_{<t},\mathbf{X},\theta)=softmax(\mathbf{W}_{o}\mathbf{F}_{\theta}(\mathbf{X})),
\end{equation}
where $\mathbf{W}_o$ is the parameter matrix for the last hidden layer, $\hat{\mathbf{Y}}_{<t}=\{y_1,...,y_{t-1}\}$. We leave the benchmarks description of medical LLM in the Sec.~\ref{sec-benchmarks}. 

\subsection{Mixture of Experts}
In modern LLMs, the mixture of experts (MoEs) strategy will significantly scale up the model size without the explosion of computational cost when deploying a LLM. Specifically, the MoE layer consists of a series of $n$ expert networks $e_1$,...,$e_n$, as well as a
gating network $G(\cdot)$ to control the model preference on those well-defined experts. Usually, we select independent MLPs to form our experts. We also utilize another lightweight MLP to serve as the gating network that assigns tokens to experts based on learned weights. For the given input token embeddings $\mathbf{x}$, the routing probability $g_i(\mathbf{x})$ for expert $i$ is formulated as:
\begin{equation}
    g_i(\mathbf{x})=softmax(\mathbf{W}_r\mathbf{x}+\epsilon)[i],
\end{equation}
where $\epsilon$ is the crafted random noise dependent on the determined routing mechanism to control the load balancing. After that, only the top-$k$ experts are activated per token during training, i.e.,
\begin{equation}
    \mathbf{H}_{route}=\sum_{i\in top_k(g(\mathbf{x}))}g_i(\mathbf{x})\cdot e_i.
\end{equation}

\section{Methodology}
\label{sec-method}
\subsection{Overview}
\label{sec-overview}

\begin{figure*}[h]
	\centering
	\includegraphics[width=0.95\textwidth,height=10.cm]{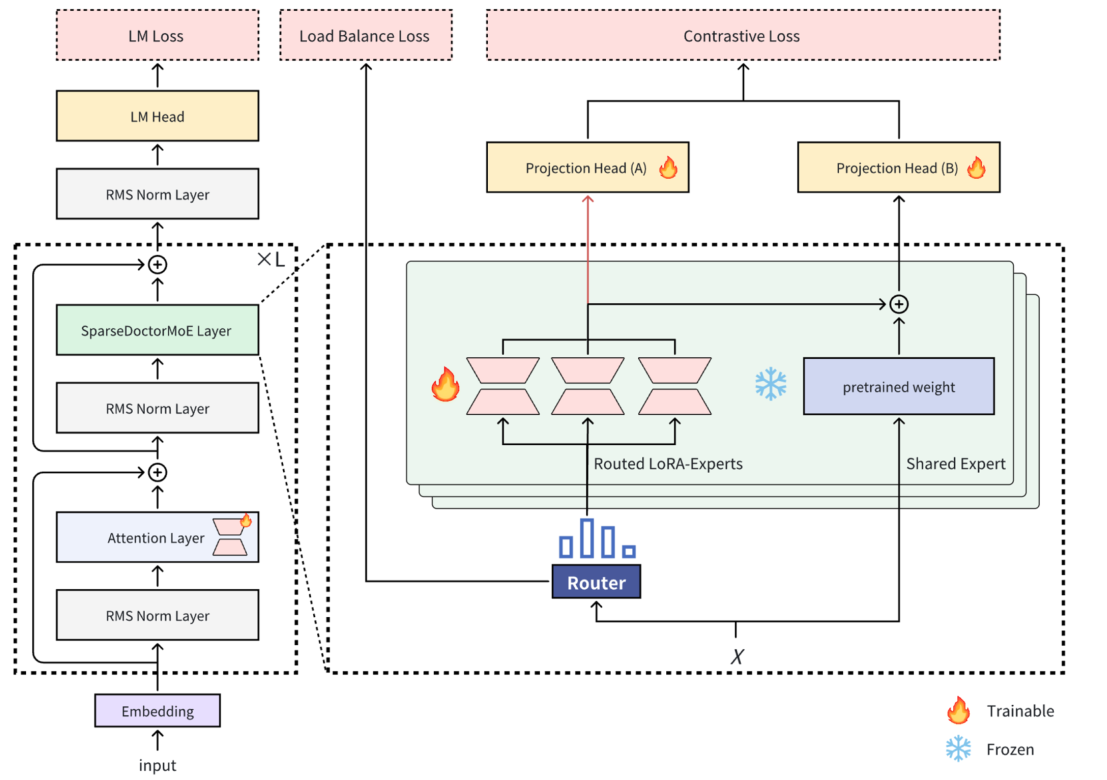}
	\caption{Overall architecture of SparseDoctor. Fire icons indicate trainable parameters, snowflake icons indicate frozen weights. The
left side shows standard Transformer layers, the right side shows the core MoE layer with shared expert, LoRA routing experts, and dual
projection head contrastive learning mechanism.}
    \label{fig-overview}
\end{figure*}

The goal of our proposed medical LLM is to increase the efficiency of the traditional medical LLM through tackling the three problems mentioned in Sec.~\ref{sec-intro}. Concretely, the method introduces multiple LoRA experts and adaptively selects the top-$k$ routers on top of the open-source large language model, i.e., Qwen3~\cite{Qwen3}. During the fint-tuning phase, we freeze the original model weights and only train the low-rank LoRA adapters, routing network, and a small number of contrastive projection layers to further achieve parameter-efficient fine-tuning. To guide experts to separately learn distinct features, we propose an expert contrastive loss and design a
memory queue to store negative samples to improve the training stability. Finally, the proposed medical LLM can achieve a balance between high performance and high efficiency in medical multi-task learning through supervision from the hybrid loss integrated by the language modeling loss, expert contrastive loss, and load balancing loss.     

\subsection{LoRA in the Base Model}
In this part, we present the details of the base model integrated with the LoRA technique to enhance the LLM's efficiency. Specifically, our proposed method adopts the open-source large language model Qwen3-4B as the backbone of the overall framework. As shown in Fig.~\ref{fig-overview}, following the parameter-efficient fine-tuning paradigm, all the pretrained parameters are frozen to ensure that the base model's memory will not be significantly affected during the fine-tuning phase. After that, we insert multiple LoRA-based experts in parallel into every fully connected layer (MLP layer) to increase model capacity. To be concrete, let $\mathbf{X}\in\mathbb{R}^{B\times T\times d}$ denotes the input hidden states to the MoE layer, where $B$ is the batch size, $T$ is the sequence length, and $d$ is the hidden dimension. We denote $e_i$ as the $i$-th expert and adopt the LoRA to encode the output latent representations of each expert into a series of paired low-rank matrices, i.e, $\mathbf{A}_i\in\mathbb{R}^{d\times r}$, $\mathbf{B}_i\in\mathbb{R}^{r\times d}$. Then, the output features of each expert $e_i$ is defined as:
\begin{equation}
    e_i=\frac{\alpha}{\tau}\mathbf{B}_i\mathbf{A}_i\mathbf{X}.
\end{equation}
It is worth noting that, unlike the conventional LoRA, we adopt the independent LoRA-based experts, where each of them learns a different representational subspace within the domain to expand the representation capability of the proposed LLM system. Next, we utilize a linear router $G: \mathbb{R}^{d}\rightarrow \mathbb{R}^{n}$ and normalize the top-$k$ experts' scores to measure the relative importance between each LoRA-based expert. Then, those normalized scores form a weight distribution $G_i(\mathbf{X})\in \mathbb{R}^{B\times T}$. After that, we obtain the fused embeddings guided by the linear router $G$:
\begin{equation}
    \mathbf{H}_{route}=\sum_{i=1}^{n}G_i(\mathbf{X})\odot e_i\in\mathbb{R}^{B\times T\times d},
\end{equation}
where $\odot$ is the element-wise Hadamard product. 

On the other hand, we also craft a parallel architecture where the frozen MLP serves as the shared expert, i.e., $\mathbf{H}_{shared}\in\mathbb{R}^{B\times T\times d}$. In the sequel, we obtain the final MoE output through the direct addition of shared and routed
expert outputs: 
\begin{equation}
    \mathbf{H}_{final}=\mathbf{H}_{shared}+\mathbf{H}_{route}.
\end{equation}
This design is similar to existing LoRA-MoE-based methods~\cite{MOELoRA, mixlora, lorahub, loramoe, MoCLE}, where the majority of them choose to insert the LoRA experts while freezing the weights of the pretrained LLMs and achieve the sparse information flow through a top-$k$ router mechanism, with the residual connections to maintain the training stability.

In the meanwhile, apart from the MLP layer, we also randomly inject LoRA adapters into the self-attention module to enhance the LLM's contextual learning capability. Specifically, after applying LoRA, the projection matrices in the attention module, i.e., $\mathbf{W}^q$, $\mathbf{W}^k$, $\mathbf{W}^v$, $\mathbf{W}^o$ are reformulated as:
\begin{equation}
    \mathbf{W}^{\prime(*)}_{attn}=\mathbf{W}^{(*)}+\frac{\alpha_{attn}}{r_{attn}}\mathbf{B}_{attn}^{(*)}\mathbf{A}_{attn}^{(*)},
\end{equation}
where $(*)\in\{q,k,v,o\}$, which represents query, key, value, and output projection matrices, respectively. $\mathbf{A}_{attn}^{(*)}$ and $\mathbf{B}_{attn}^{(*)}$ are the low-rank decompositions accordingly. 

\begin{figure*}[h]
	\centering
	\includegraphics[width=0.95\textwidth,height=10.cm]{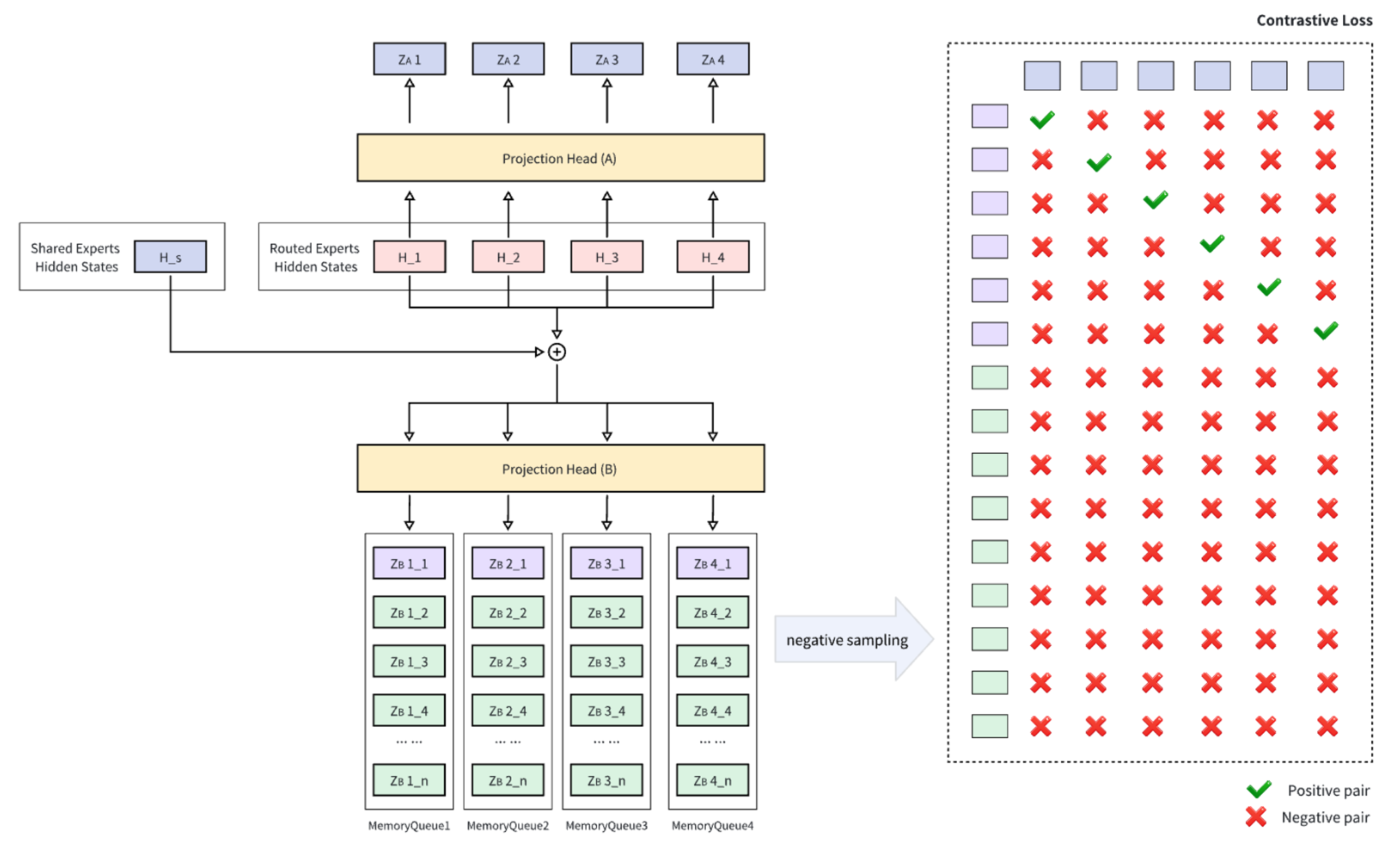}
	\caption{Detailed flow of expert contrastive learning mechanism. The left diagram shows the dual projection head architecture and memory queue mechanism, while the right diagram shows the positive and negative sample pair construction strategy for InfoNCE contrastive loss. Green checkmarks indicate positive pairs, red crosses indicate negative pairs.}
    \label{fig-contrastive-overview}
\end{figure*}

\subsection{Load Balancing in MoE}
During training, we will easily encounter the ``OOM" issue due to the contradiction between the vast parameter space of the LLM and the limited computational resources. To alleviate this problem, we endeavor to craft a load-balancing mechanism through an information-theoretical perspective. Specifically, we craft a KL divergence~\cite{KLD} to push the routing distribution toward a uniform distribution:
\begin{equation}
    \mathcal{L}_{balance}=KL(Uniform(n)||\mathbf{\bar{P}})=\sum_{i=1}^{n}\frac{1}{n}\log\frac{1/n}{\bar{\mathbf{P}}_{i}},
\end{equation}
where $\mathbf{\bar{P}}_i=\frac{1}{BT}\sum_{b,t}G_i(\mathbf{X}_{b,t})$ is the average routing probability of expert $i$ across all positions, and $n$ is the total number of experts. Under these circumstances, we could evenly distribute the computation among all the experts via the near-uniform routing distribution. 

\subsection{Expert-Based Contrastive Learning}
In this part, we provide an in-depth analysis of the limitations of the vanilla MoE framework and introduce a novel contrastive learning framework to supervise the routing mechanism of MoE. The vanilla MoE frameworks suffer from two critical issues. First, the ``load balancing problem"~\cite{AcceLLM} leads to a phenomenon that a few "hot" experts dominate the routing stage while the others remain idle, which will cause a significant wasting of computational resources.   
Second, even when the load is balanced, the gating network~\cite{gating_network} often shows no preference for specific experts and will lead to the random routing issue, i.e., the tokens that are distributed to different experts will share similar content and will push all the experts to learn nearly identical representations. The prior work~\cite{MOELoRA} mitigates this by treating outputs from the same expert as positive examples and those from different experts as negative ones. However, constructing cross-token positive pairs requires the gating
network to be reliable from the outset, and incurs significant memory overhead when scaling to long sequences. 

To address this issue, we propose an alternative expert contrastive loss that circumvents cross-token pairing by constructing two complementary views per
token and applying the noise-contrastive estimation.  
To capture complementary information, we apply two lightweight projection heads in parallel to generate the dual views required for contrastive learning. Specially, we first generate the \textit{routed expert view} (\textbf{view A}):
\begin{equation}
    \mathbf{Z}^{\mathbf{A}}=\mathbf{W}_{2}^{\mathbf{A}}ReLU(\mathbf{W}_{1}^{\mathbf{A}}Drop(\mathbf{H}_{route})),
\end{equation}
where $Drop(\cdot)$ indicates the dropout function~\cite{dropout}, which aims to introduce randomness into the feature space to further enhance the LLM's representation capability; $\mathbf{W}_{1}^{\mathbf{A}}$, $\mathbf{W}_{2}^{\mathbf{A}}$ are the projection matrices in the first view. On the other hand, we introduce the \textit{fused expert view} (\textbf{view B}) as its counterpart to align the feature engineering of the different experts that share the same tasks and tell apart the representations obtained from different experts that have distinct tasks. Specially, the output embeddings generated by the fused expert view are formulated as:
\begin{equation}
    \mathbf{Z}^{\mathbf{B}}=\mathbf{W}_{2}^{\mathbf{B}}ReLU(\mathbf{W}_{1}^{\mathbf{B}}Drop(\mathbf{H}_{route}+\lambda\mathbf{H}_{shared})),
\end{equation}
where $\mathbf{W}_{1}^{\mathbf{B}}$ and $\mathbf{W}_{2}^{\mathbf{B}}$ are projection matrices of the fused expert view accordingly, $\lambda$ is the hyperparameter to control the relative importance of the shared expert information in the fusion view. It is worth noting that following the core idea of unsupervised SimCSE~\cite{simcse}, we apply the independent dropout masks to create two slightly different embeddings of the same token to avoid the representation collapse~\cite{mode_collapse}. However, unlike SimCSE, our method requires no repeated forward passes but constructs semantically related yet feature-distinct contrastive views through residual fusion.

In fact, we will encounter memory overflow issues in large-scale contrastive learning (the sample size is too large in the LLM field). We then novelly craft an expert memory queue mechanism to alleviate this problem during training. We first introduce the design of the expert memory queue. Each expert maintains a fixed-length circular queue storing the view B's projection vectors in historical steps, i.e., $\textbf{MemoryQueue}_{i}\in\mathbb{R}^{K\times d_h}, i=1,...,n$. Then, for each batch, we formulate the determination of the highest activated expert for each token as an optimization problem:
\begin{equation}
    \textbf{MemoryQueue}_i[ptr_i]=\mathbf{Z}_{b,t}^{B}, \ \text{if } i=\arg\max_{j} G_j(\mathbf{X}_{b,t}), 
\end{equation}
where $ptr_i$ is the circular write pointer for expert $i$, $\mathbf{Z}_{b,t}^{B}$ is the corresponding output of the memory queue, and is enqueued into expert $i$'s circular buffer. When the queue is full, new features overwrite the oldest ones. Under this scenario, the memory complexity will be reduced from $\mathcal{O}(B\times T \times N)$ to $\mathcal{O}(K \times N)$ and can greatly prevent the OOM issues in large-scale training. After that, we flatten $\mathbf{Z}^{\mathbf{A}}$ and $\mathbf{Z}^{\mathbf{B}}$ into $\mathbf{z}_a\in\mathbb{R}^{N\times d_h}$ and $\mathbf{z}_b\in\mathbb{R}^{N\times d_h}$, where $N=B\times T$. We let $\mathcal{Q}=\{q_1,...,q_M\}$ be a set of $M$ negative features sampled uniformly from all queues. We adopt the InfoNCE objective~\cite{infoNCE} to bring each positive pair $(z_{a,i},z_{b,i})$ closer while pushing the negative pairs away from each other: 
\begin{equation}
    \mathcal{L}_{co}=\frac{1}{N}\sum_{i=1}^{N}-\log\frac{exp\{z_{a,i}^{\top}z_{b,i}/\tau\}}{exp\{z_{a,i}^{\top}z_{b,i}/\tau\}+\sum_{j=1}^{M}exp\{z_{a,i}^{\top}q_{j}/\tau\}},
\end{equation}
where $\tau>0$ is a temperature hyer-parameter. This loss aligns two complementary views on the same token while encouraging distinction from
negative samples drawn across all experts. Importantly, it does not require the router to generate meaningful token-to-expert assignments
during the early training phase because the positive pairs are constructed from the same token. As training progresses, the negative queue encourages the experts to consistently regard the samples from different experts as negatives and further promote the separation between their representations. 

\subsection{Training Loss}
The entire proposed medical LLM only trains LoRA experts, routers, and contrastive projection layers, while all other original model weights are completely frozen. In this scenario, the proposed medical LLM could easily be fine-tuned from other high-quality pre-trained open-source LLMs and pave the way to a powerful and resource-saving medical LLM. To fine-tune the parameters in the adjustable modules of the proposed LLM, we formulate the loss function as the weighted summation of the language modeling loss, load-balancing loss, and contrastive loss:
\begin{equation}
    \mathcal{L}_{total}=\mathcal{L}_{LLM}+\alpha\mathcal{L}_{balance}+\beta\mathcal{L}_{co},
\end{equation}
where $\alpha$ and $\beta$ are the hyperparameters to adjust the relative importance of load-balancing and contrastive losses during training.

\section{Experiments}
\label{sec-exp}
In this section, we first introduce the details of the datasets with their statistics in Sec .~\ref {sec-dataset}. We then introduce the details of the corresponding experiment settings, including the hyperparameter setting, the model architecture, and the training strategies in Sec .~\ref {sec-exp-setting}. Later, we give a comprehensive description of the baseline models and benchmarks on the medical question answering task in Sec .~\ref {sec-benchmarks}. We then provide extensive experiments in the rest of Sec .~\ref {sec-exp} to answer the following three questions:
\begin{itemize}
    \item \textbf{RQ1}: How does the proposed method perform on the medical question answering task compared to other LLMs?
    \item \textbf{RQ2}: How do different modules impact model performance?
    \item \textbf{RQ3}: What are the influences of varying hyperparameters?
\end{itemize}

\subsection{Dataset Description}
\label{sec-dataset}
To evaluate the efficient adaptation in the Chinese medical domain of our proposed method, we construct a dedicated dataset based on a publicly
available large-scale Chinese medical instruction corpus, comprising $100k$ training samples and $5k$ validation samples. The dataset encompasses nine distinct medical knowledge sources, which adopt a fixed source-quota with within-source random sampling strategy, together with a unified instruction-tuning triplet format, response-span supervision, and length control, ensuring a stable and reproducible data distribution without highly relying on complex curriculum learning or priority scheduling. This subsection details the dataset construction process, including source provenance, data pre-processing strategies, sampling methods, and the final dataset configuration.

The dataset comprises $100000$ training samples and $5000$ validation samples in Chinese. Most of them are obtained through quota-based, intrasource, and equal-probability random sampling strategy from a public medical corpus (without any priority or curriculum sorting strategies; samples are randomly shuffled within each training epoch). The quotas, counts, and proportions of nine data sources are as follows (validation set
allocated as $5\%$ ratio):

\begin{table}[h]
	\centering
	\caption{Dataset statistics.}
	\label{tab-dataset}
	\resizebox{0.8\columnwidth}{!}{%
		\begin{tabular}{c|ccc}
			\toprule[1.pt]
			Source & \#Train & \#Val & Train ($\%$) \\
			\hline
			knowledge\_Web\_Corpus\_en & 7000  & 350 & 7.0 \\ 
			knowledge\_Web\_Corpus\_cn & 12000 & 600 & 12.0\\
            knowledge\_Literature\_en & 16000 & 800 & 16.0\\
            knowledge\_Literature\_cn & 3000  & 150 & 3.0\\
          knowledge\_Encyclopedia\_en & 3000  & 150 & 3.0\\
          knowledge\_Encyclopedia\_cn & 8000  & 400 & 8.0\\
                 knowledge\_Books\_en & 15000 & 750 & 15.0\\
                 knowledge\_Books\_cn & 34000 & 1700& 34.0\\
                           dialogue & 2000  & 100 & 2.0\\
			\bottomrule[1.pt]
		\end{tabular}
	}
\end{table}

It is worth noting that "\_en/\_cn" suffixes in source names only identify the source channels (English/Chinese sources), not the language of the samples. Moreover, all $105k$ samples obtained in this study are Chinese Q\&A dialogues (consistent with the source dataset construction approach of "setting target language to Chinese").

To ensure medical coverage and knowledge density, we sample from the large-scale Chinese medical instruction corpus constructed by HuatuoGPT-II~\cite{HuatuoGPTII}. Specifically, this corpus initially aggregated approximately $1.1TB$ of medical-related raw text from sources incorporating encyclopedias, books, papers, and web pages across general and specialized channels. We also adopt a series of complex strategies, including medical vocabulary filtering, sentence segmentation, advertisement/noise filtering, and semantic deduplication processes to obtain the approximately $5.25M$ medical corpus entries. We then uniformly convert those entries to an instruction format, i.e., question-answer pairs for subsequent training. Additionally, the corpus incorporates approximately $140K$ high-quality medical Q\&A samples, consisting of real medical questions paired with professional answers generated by GPT-4~\cite{gpt4}, where questions are derived from Huatuo-$26M$ (Chinese Medical Q\&A Collection).  

\subsection{Experiment Settings}
\label{sec-exp-setting}
We adopt Qwen3-4B as the base model with the standard transformer~\cite{transformer} configurations (hidden size equals $2560$, intermediate size equals $9728$, $36$ layers, $32$ attention heads, maximum positional embeddings equals $40,960$, and a SiLU activation function~\cite{SiLU}). 
All pre-trained weights are frozen except for the adapters, the router, and the contrastive projection heads to ensure the parameter-efficient transfer without increasing inference overhead. 
A sparse MoE-LoRA structure is introduced in the MLP sublayer~\cite{MLP}: each layer hosts $16$ parallel LoRA experts (LoRA rank equals $16$, scaling factor alpha equals $32$). A top-$k$ router activates $4$ experts per token and fuses their outputs using softmax-normalized weights. The shared (frozen) MLP output and the routed experts' output are added via a residual connection. This design follows LoRA's parameter-efficient paradigm of "frozen backbone plus low-rank increments" together with sparse gating, which can further expand model capacity and improve multi-task adaptability without significantly increasing computational complexity. As for the self-attention module, the standard LoRA adapters are injected into the query, key, value, and output projection matrices, i.e., $q\_proj, k\_proj, v\_proj$, and $o\_proj$ accordingly, to enhance sequence modeling without altering the computational graph of attention. 

To mitigate random routing and expert representation convergence issues, we attach dual projection heads to the MoE output of each layer and formulate an InfoNCE-based expert contrastive objective, i.e., View A operates directly on the routed experts' features, while View B fuses the shared expert features with the routed features. The temperature is set to $0.07$, and each expert maintains a ring buffer queue of length equals $8$ to provide stable negative samples. This approach is inspired by SimCSE's strategy of using "dropout views as minimal data augmentation" to avoid representation collapse, while not relying on accurate early routing. For auxiliary losses, both the KL divergence term for routing load balancing and the expert contrastive loss are weighted at $0.01$, while the other hyperparameters follow the base model defaults.

The training objective is autoregressive language modeling with a multi-objective joint optimization that combines the language modeling loss, the load-balancing loss, and the expert contrastive loss. We use the AdamW optimizer~\cite{AdamW} with a base learning rate of $1\times 10^{-4}$. After a $200$-step linear warm-up, a cosine annealing schedule~\cite{adaptivelearningrate} is applied (the minimum learning rate is one tenth of the base rate). The global gradient clipping threshold is $1.0$. The per-device batch size is $8$ with $2$ gradient accumulation steps. We use $bf16$ mixed precision during the training phase to further decrease the computational overhead.  

For validation, we compute the loss every $100$ steps on $5000$ stratified samples for early monitoring. Instruction-tuning examples follow a unified ChatML format, and the supervision is applied only to the answer span. Moreover, the maximum input sequence length (tokenized prompt plus response) is set to $512$. These settings conform to standard practices for parameter-efficient fine-tuning of LLMs~\cite{lora, lora+, lorahub, loramoe} and stable training with sparse routing.

We evaluate on three public Chinese medical benchmarks: CMB (Comprehensive Medical Benchmark in Chinese), CMExam (derived from
China's National Medical Licensing Examination), and CMMLU-Med (the medical subset of CMMLU). All objective-type questions are
measured by accuracy. 
The evaluation script extracts answer options from generated text via rule-based matching and reports both overall performance and per-subset results. During the inference phase, we use deterministic greedy decoding (do\_sample=False, num\_beams=1), with a maximum of $512$ new tokens in left padding (padding\_side='left') manner, and the ChatML prompt template to avoid randomness from temperature and sampling. We conduct 5 individual repeated experiments and report the corresponding mean values.  

\subsection{Baselines \& Benchmarks}
\label{sec-benchmarks}
To comprehensively evaluate the conversational ability of the proposed model in Chinese medical scenarios, this study selects six representative models in Chinese medical or bilingual (Chinese–English) dialogue as baselines, covering medical-domain models, general-purpose dialogue models, and a closed-source upper-bound reference. The models' training characteristics and openness are listed below:

\begin{itemize}
    \item \textbf{Baichuan2-7B-Chat}~\cite{baichuan2}: The Baichuan2 series adopts a staged training strategy on large-scale, high-quality corpora. According to the official technical report, the 7B base model is trained on approximately 2.6 trillion tokens and further yields a Chat variant tailored for dialogue tasks; the series releases intermediate checkpoints to facilitate academic research and model analysis.
    \item \textbf{ChatGLM3-6B}~\cite{chatglm}: ChatGLM3 is the latest generation of the GLM family for bilingual (Chinese–English) dialogue. It provides standard dialogue weights, a base version, and a long-context (32K) version. Model weights are fully open for academic use and permitted for commercial applications, with solid engineering support and ecosystem readiness.
    \item \textbf{GPT-4}~\cite{gpt4}: As a closed-source upper-bound reference, GPT-4 is a multimodal large model that demonstrates strong performance on diverse professional and academic benchmarks in its technical report. Because its training details are not public, we use it only as a reference point in the literature comparison.
    \item \textbf{DISC-MedLLM}~\cite{DISC-MedLLM}: A Chinese medical large language model designed for medical dialogue. It constructs high-quality supervised fine-tuning (SFT) data via three strategies—medical knowledge graphs, reconstruction of real doctor–patient dialogues, and human preference rewriting—and publicly releases approximately 470,000 training instances with an open-source implementation, emphasizing practicality and robustness for real-world medical consultations. 
    \item \textbf{HuatuoGPT}~\cite{huatuogpt}: A medical-dialogue LLM trained with a mixed-data strategy that combines ChatGPT-distilled answers and physician-annotated data during SFT. The study analyzes the properties and complementarity of these two data sources and designs the training recipe accordingly to improve clinical question-answering utility.
    \item \textbf{HuatuoGPT-II}~\cite{HuatuoGPTII}: This model proposes a unified single-stage medical-domain adaptation method that integrates continued pretraining and supervised fine-tuning into one pipeline, and introduces a priority-based sampling strategy to mitigate distribution shift and catastrophic forgetting. It reports systematic results on multiple Chinese medical benchmarks and serves as an important baseline reference for Chinese medical evaluation. The baseline performance figures used in this paper are taken from its published results and cross-checked against the original sources.
\end{itemize}

The evaluation data cover typical application scenarios such as multiple-choice questions in Chinese medical licensure style and clinical inquiry. All benchmarks are standard, publicly accessible resources that support third-party verification and replication. All baseline comparisons use accuracy as the primary evaluation metric and strictly follow each dataset's official splits and canonical formatting.
\begin{itemize}
    \item \textbf{CMB (Chinese Medical Benchmark)}: CMB is designed to systematically assess Chinese medical knowledge and reasoning, and comprises exam-style and clinical subsets. The CMB-Exam subset is organized into $6$ primary categories and $28$ subcategories, with $400$ questions per subcategory and $11200$ questions in the test set; CMB-Clin contains $74$ complex clinical case inquiries. The official repository and dataset card clearly document composition and intended use.
    \item \textbf{CMExam}: An objective-question benchmark derived from Chinese medical licensure/qualification examinations. The paper describes the construction pipeline, annotation dimensions, and a unified scoring protocol, while the dataset card provides download and citation information. CMExam is widely used to assess coverage of Chinese medical knowledge points and problem-solving ability.
    \item \textbf{CMMLU (medical subset)}: CMMLU is a large-scale Chinese multi-discipline benchmark covering $67$ subjects, formatted as single-answer multiple-choice (four options). Its medical subset focuses on clinical medicine, pharmacy, and biomedical fields. Official scripts and the dataset card provide standardized loading and evaluation procedures.
\end{itemize}

We evaluate on three Chinese medical benchmarks, i.e., CMB-Exam, CMExam, and CMMLU-Med, using a unified greedy decoding configuration (do\_sample=False, num\_beams=1). The maximum number of newly generated tokens is set to $512$. All multiple-choice tasks are scored by predicting accuracy. To reflect the relative importance of each benchmark, the overall score is computed as the weighted average of the three benchmarks: let the numbers of questions be $N_{B}$, $N_{E}$, $N_{U}$. ($11200$; $6811$; and $1354$ respectively), with corresponding accuracies $Acc_{B}$, $Acc_{E}$, $Acc_{U}$. The weighted average is formulated as:

\begin{equation}
    Average=\frac{N_{B}\cdot Acc_{B}+N_{E}\cdot Acc_{E}+N_{U}\cdot Acc_{U}}{N_{B}+N_{E}+N_{U}}
\end{equation}

Based on the above evaluation setup, the main results are summarized in Tab.~\ref{tab-exp1}. 

\subsection{Experiment Results}
Tab.~\ref{tab-exp1} presents the performances of the baselines with our proposed medical LLM on the clinical-related dataset. Specifically, we report the corresponding performances based on the aforementioned three benchmarks and the crafted average score to comprehensively measure the model comparison results on the medical question answering task. It is worth noting that ``$\Delta$ vs. HuatuoGPT-II" denotes the performance gap between ours and HuatuoGPT-II.  

\begin{table}[h]
	\centering
	\caption{Main results on Chinese medical benchmarks (Accuracy, $\%$).}
	\label{tab-exp1}
	\resizebox{0.8\columnwidth}{!}{%
		\begin{tabular}{c|cccc}
			\toprule[1.pt]
			Model & CMB & CMExam & CMMLU-Med & Average \\
			\hline
			DISC-MedLLM & 32.47 & 36.62 & –    & – \\
            HuatuoGPT   & 28.81 & 31.07 & 33.23& 29.91 \\
            ChatGLM3 & 39.81 & 43.21 & 46.97& 41.51 \\
       Baichuan2& 46.33 & 50.48 & 50.74& 48.10 \\
          GPT-4       & 43.26 & 46.51 & 50.37& 44.90 \\
            HuatuoGPT-II& 60.39 & 65.81 & 59.08& 62.20 \\
            Qwen3    & 60.87 & 63.66 & 65.81& 62.19 \\
            SparseDoctor& \textbf{62.54} & \textbf{66.88} & \textbf{68.54}& \textbf{64.49} \\
        \hline
      $\Delta$ vs. HuatuoGPT-II & +2.15 & +1.07 & +9.46& +2.29 \\
      $\Delta$ vs. Qwen3     & +1.67 & +3.22 & +2.73& +2.30 \\
			\bottomrule[1.pt]
		\end{tabular}
	}
\end{table}

Compared to the latest open-source baseline HuatuoGPT-II, SparseDoctor improves the average score by $2.29\%$, which demonstrates that our contrastive learning enhanced LoRA-MoE strategy can significantly improve the comprehension and inference capability of the LLMs on the medical domain. We also introduce the original backbone model Qwen3~\cite{Qwen3} in the baselines list to disentangle backbone strength from SparseDoctor. The corresponding experiment results indicate that SparseDoctor still provides a $2.30\%$ net gain, demonstrating that the improvement primarily arises from the proposed contrastive learning enhanced MoE-LoRA architecture. On the other hand, based on the CMMLU-Med metric, SparseDoctor outperforms HuatuoGPT-II by $9.46\%$ and Qwen3 by $2.73\%$, demonstrating that the backbone model contributes roughly $6.73\%$ performance gain on the medical Q\&A task, while the remaining $2.73\%$ net gain comes from the contrastive learning enhanced LoRA-MoE design. This phenomenon reflects that, apart from the vanilla data perspective insight~\cite{huatuogpt,HuatuoGPTII,chatdoctor,doctorglm}, the proposed architecture-based improvements can significantly promote expert specialization and mitigate inter-task interference, thereby enhancing consistency in cross-disciplinary medical reasoning.

\subsection{Ablation Studies}

To analyze the contribution of each module in SparseDoctor individually, we conduct a \textit{stepwise ablation studies} under identical training and inference settings. Starting from the backbone Qwen3, we sequentially add MLP-MoE (LoRA experts), LoRA on the attention projections (Q/K/V/O), and expert contrastive learning (dual projections plus InfoNCE plus expert memory queues), ultimately forming the proposed SparseDoctor. Except for the module under consideration, all other hyperparameters are kept consistent with Sec .~\ref {sec-method}. 

We evaluate four experiment groups: ``Backbone", ``+MLP-MoE", ``+Attn-LoRA", and ``+Contrast". The design motivations align with recent LoRA-MoE literature, i.e., the parameters of the backbone model are frozen, and the model's representation capacity is expanded via multiple LoRA experts and a router. At the same time, the contrastive learning module is used to suppress the convergence of expert representations. Tab.~\ref{tab-ablation} summarizes the performances across all experimental groups, which presents the progressive improvement as components are added sequentially. The results indicate that the three newly introduced modules can boost the performance of the medical LLM to some extent. It is noteworthy that the performance gain for introducing contrastive learning is the largest, i.e, $0.97\%$, which indicates that our contrastive learning module plays a vital role in boosting the knowledge comprehension and reasoning capability of the LLM on the medical domain.     

\begin{table}[h]
	\centering
	\caption{Ablation results (Accuracy, \%).}
	\label{tab-ablation}
	\resizebox{0.8\columnwidth}{!}{%
		\begin{tabular}{c|cccc}
			\toprule[1.pt]
			Variant & CMB & CMExam & CMMLU-Med & Average \\
			\hline
			Backbone   & 60.87 & 63.66 & 65.81 & 62.19 \\
            +MLP-MoE   & 60.96 & 64.56 & 66.99 & 62.64 \\
            +Attn-LoRA & 61.56 & 65.79 & 67.21 & 63.44 \\
            +Contrast (final model) & 62.37 & 66.94 & 68.61 & 64.41 \\
			\bottomrule[1.pt]
		\end{tabular}
	}
\end{table}

\begin{figure*}[h]
	\begin{center}
        \subfloat[Projection Dimension]{\includegraphics[width=0.24\textwidth,height=2.3cm]{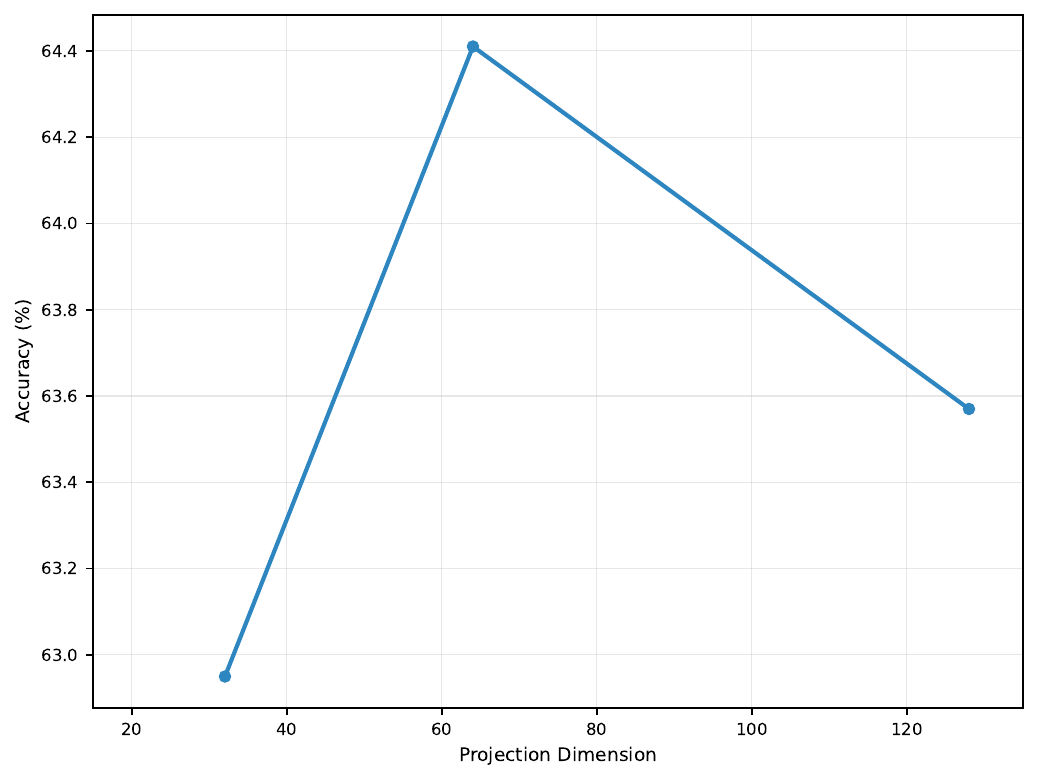}}
        \hfill
		\subfloat[Queue Size]{\includegraphics[width=0.24\textwidth,height=2.3cm]{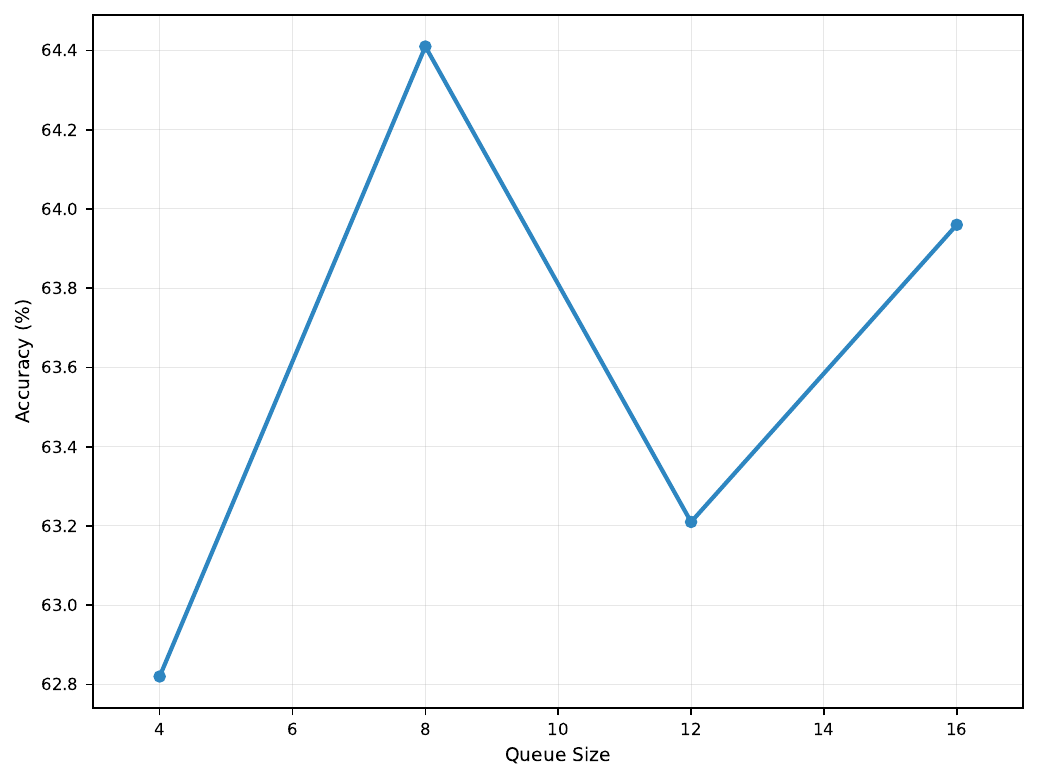}}
        \hfill
        \subfloat[$\beta$]{\includegraphics[width=0.24\textwidth,height=2.3cm]{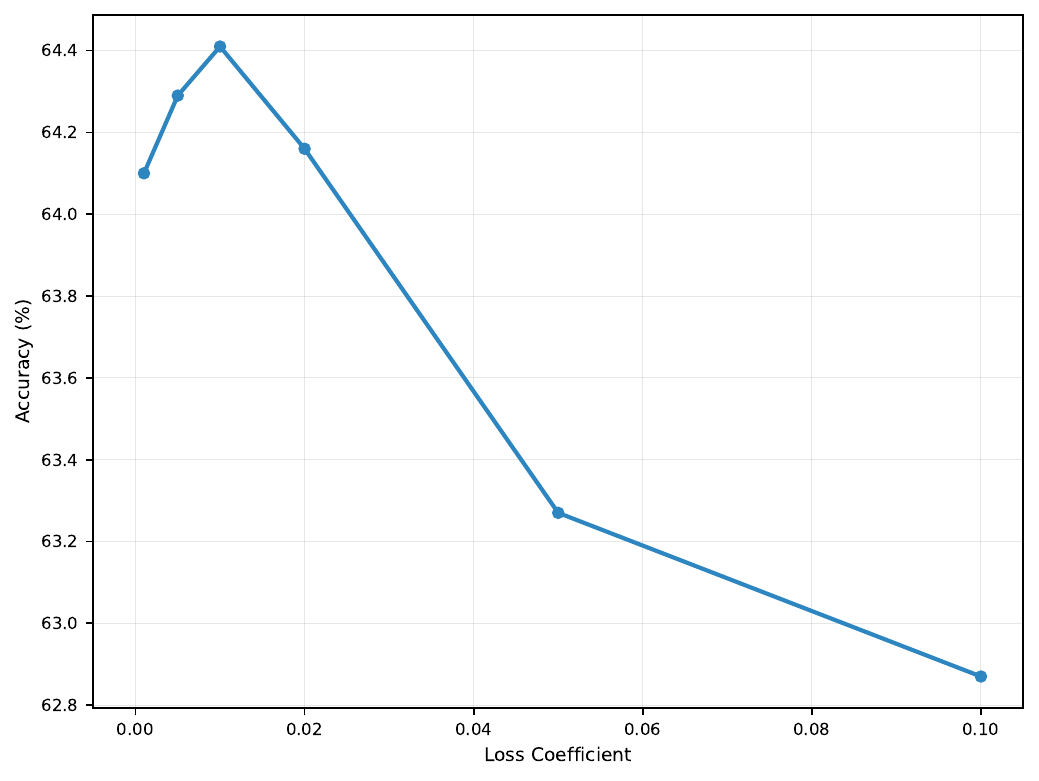}}
        \hfill
        \subfloat[Temperature]{\includegraphics[width=0.24\textwidth,height=2.3cm]{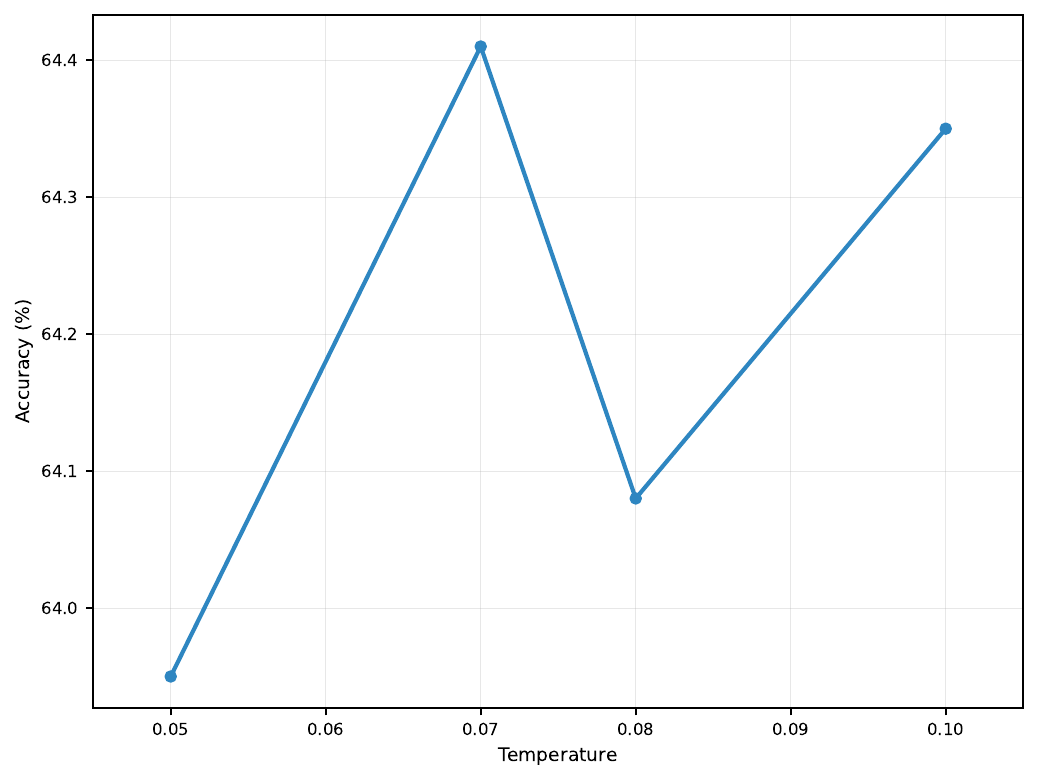}}
	\end{center}
	\caption{Sensitivity analysis.}
	\label{fig-Sensitivity}
\end{figure*}

\subsection{In-depth Analysis on Expert Contrastive Learning}

We further provide an in-depth analysis of the expert contrastive learning module to justify its importance in the LoRA-MoE architecture. To this end, we design a controlled ablation experiment focusing on its impact on \textit{routing confidence}, which is a key metric for the determinism of the router's decisions. That is, an increase in routing confidence directly reflects the mitigation of random routing.

Firstly, we construct the two model variants that are identical in architecture and training configuration: a baseline model (contrastive learning disabled) and the proposed model (expert contrastive learning enabled). The only difference lies in the composition of the loss function, i.e., the proposed model additionally introduces an InfoNCE-based expert contrastive loss and a memory queue mechanism. 
Specifically, the routing confidence is defined as the expected maximum of token-level routing weights:

\begin{equation}
    Conf=E_{x}[max_{e\in\{1,2,...,E\}} G_e(x)],
\end{equation}
where $G_e(x)\in(0,1)$ denotes the normalized weight assigned by the router to expert $e$ for input $x$, and $\sum_{e=1}^{E}G_e(x)=1$. Higher values indicate more deterministic routing decisions and fewer random routing phenomena.

We evaluate on a combined test set from the Chinese medical benchmarks, i.e., CMB, CMExam, and CMMLU-Med, which comprise $19365$ medical QA samples and around $4.81\times10^{7}$ tokens. Tab.~\ref{tab-conf} reports the impact of expert contrastive learning on global routing confidence. Under otherwise identical
conditions, introducing the expert contrastive learning mechanism increases global routing confidence from $0.3674$ to $0.3716$, yielding a $1.13\%$ relative improvement.

\begin{table}[h]
	\centering
	\caption{Impact of expert contrastive learning on global routing confidence.}
	\label{tab-conf}
	\resizebox{0.8\columnwidth}{!}{%
		\begin{tabular}{c|c|c|c}
			\toprule[1.pt]
			Metric & w.o. contrastive learning  & w/ contrastive learning & Change\\
			\hline
			Conf & $0.3674$ & $0.3716$ & $+1.13\%$ \\ 
			\bottomrule[1.pt]
		\end{tabular}
	}
\end{table}

\begin{figure*}[h]
	\centering
	\includegraphics[width=0.9\textwidth,height=10cm]{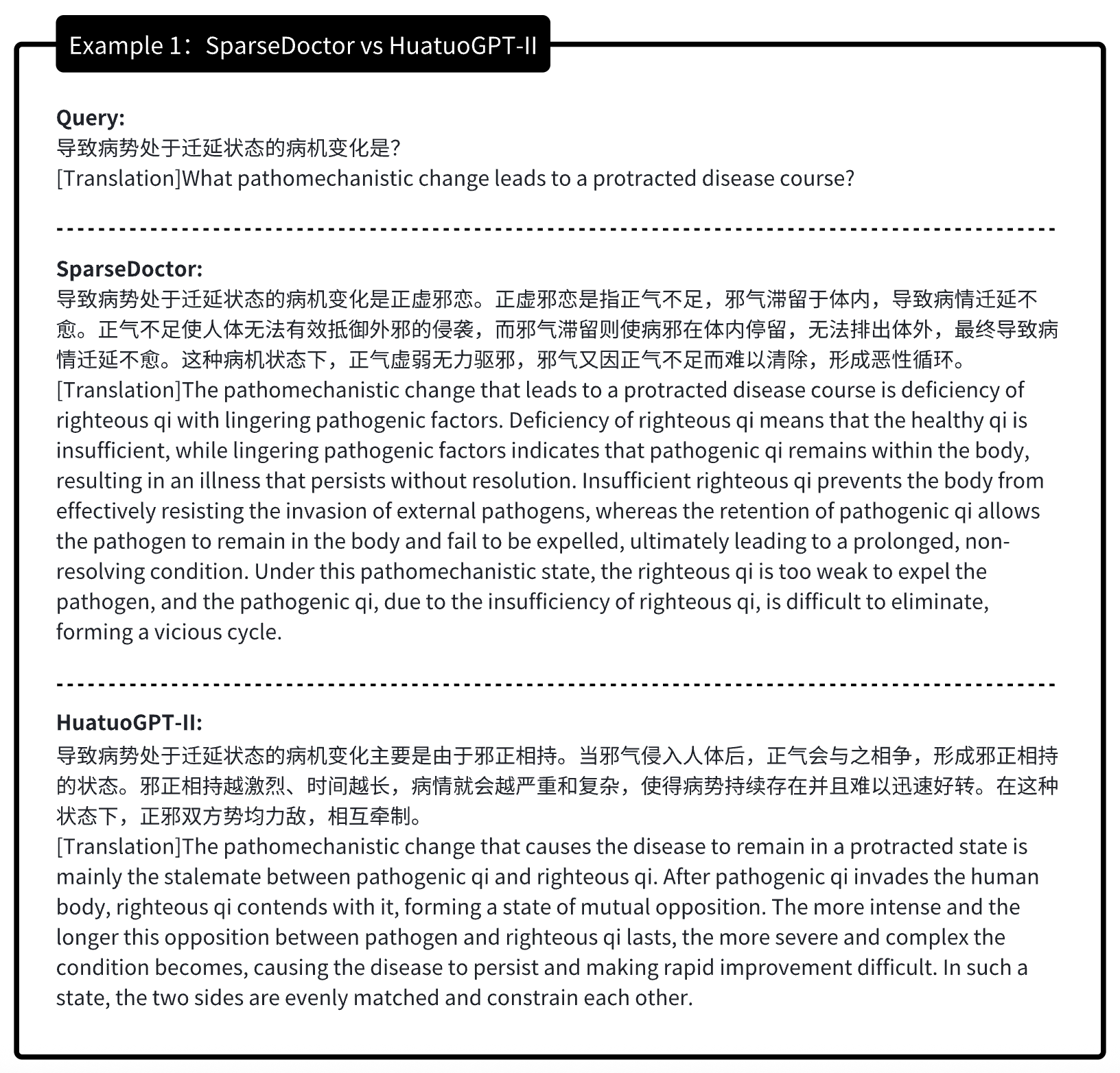}
	\caption{\textbf{Case 1}: Comparative case on TCM pathomechanism theory. It shows the difference between SparseDoctor and HuatuoGPT-II in
understanding TCM pathomechanism theory. SparseDoctor accurately identifies "zheng-xu xie-lian" as the key pathomechanism of a
protracted disease course, demonstrating a precise grasp of the core concepts of TCM theory.}
    \label{fig-case-1}
\end{figure*}

\begin{figure*}[h]
	\centering
	\includegraphics[width=0.9\textwidth,height=10cm]{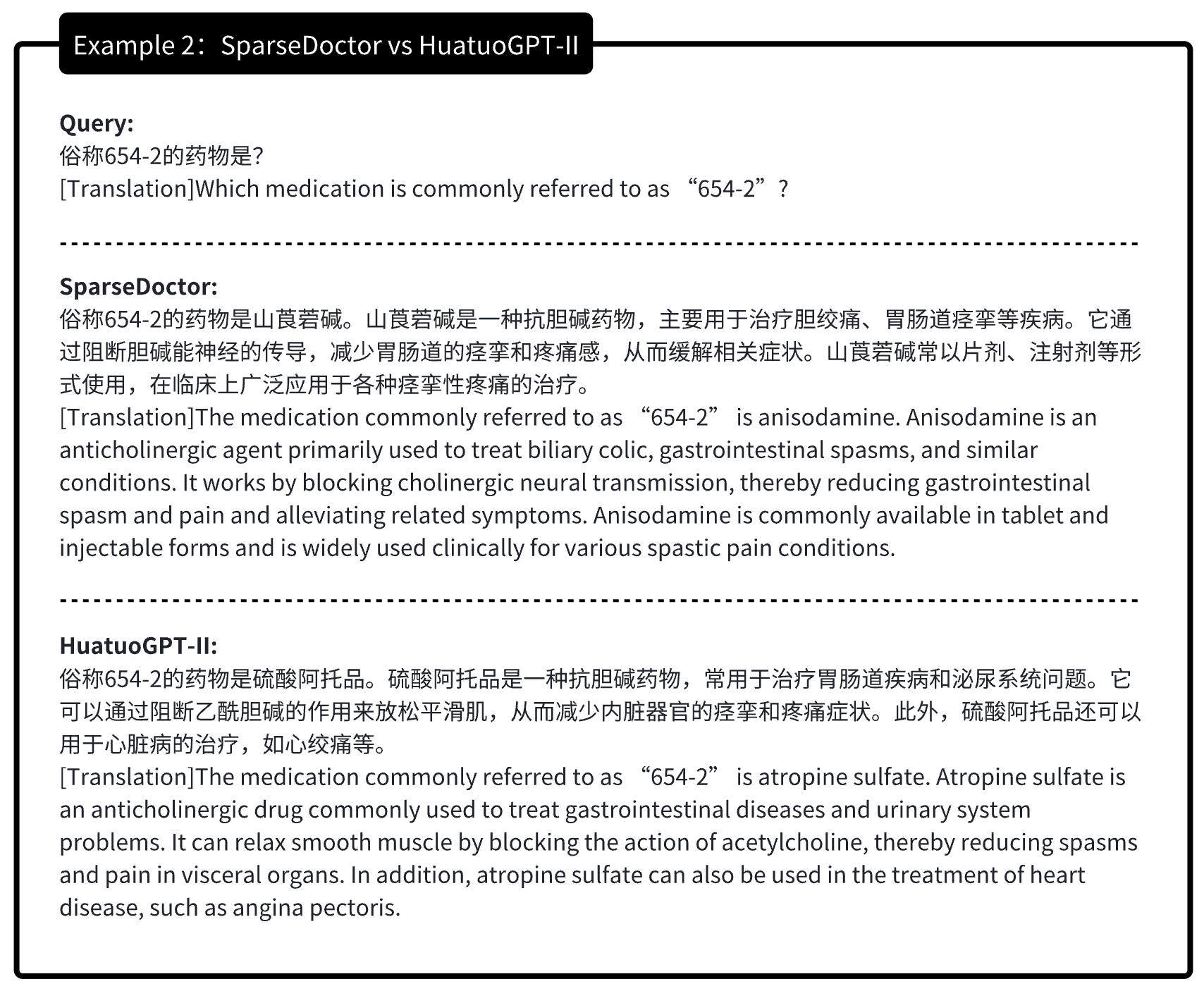}
	\caption{\textbf{Case 2}: Comparative case on drug identification knowledge. It contrasts SparseDoctor and HuatuoGPT-II in drug code
identification. SparseDoctor correctly identifies 654-2 as anisodamine, reflecting accurate mastery of the Chinese drug numbering system and precision in drug knowledge.}
    \label{fig-case-2}
\end{figure*}

\begin{figure}[h]
	\centering
	\includegraphics[width=0.8\textwidth,height=4cm]{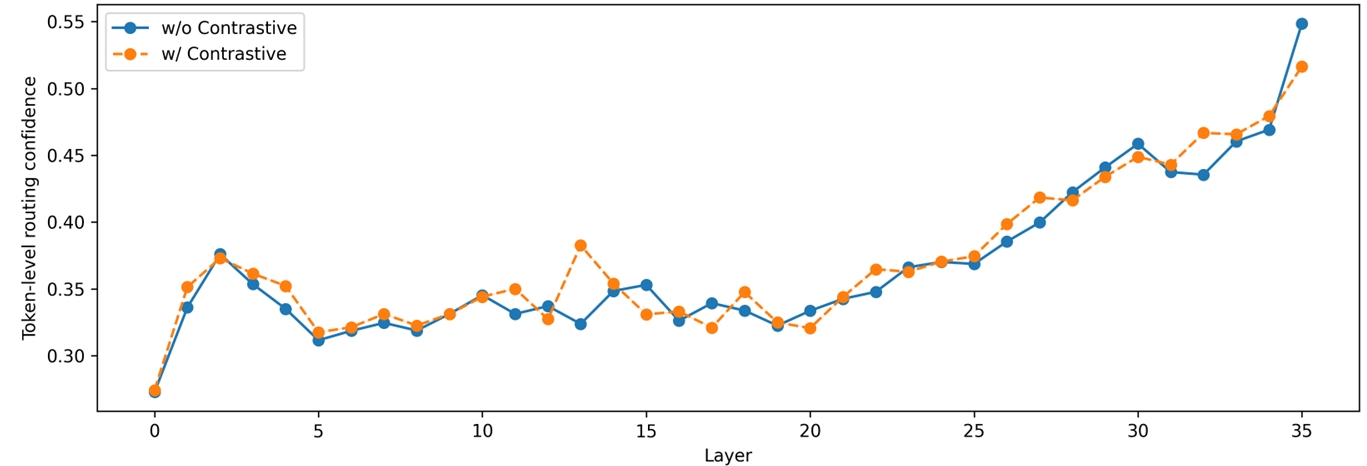}
	\caption{Routing confidence as a function of network depth. The horizontal axis denotes layer index (0–35), and the vertical axis
denotes the average routing confidence of that layer. The orange dashed line represents the model with expert contrastive learning
enabled, and the blue solid line represents the baseline model.}
    \label{fig-conf-vs-layer}
\end{figure}

To further understand how this improvement is distributed across depth, Fig.~\ref{fig-conf-vs-layer} depicts the routing confidence as a function of network
depth under the two model configurations. It is observed that by constructing a dual-view InfoNCE-based contrastive objective, the expert contrastive learning effectively alleviates the prevalent random routing issue in MoE models. Its core roles are: \textbf{1)} enhancing inter-expert representational distinctiveness and enabling the router to learn clearer expert selection patterns; \textbf{2)} providing a more stable contrastive signal via a memory queue that maintains
historical negative samples; \textbf{3)} ensuring effective decoupling between the contrastive objective and the main task objective through a dual-projection-head design.

The above experimental results validate the effectiveness of the expert contrastive learning mechanism, offering a parameter-efficient and effective solution to the random routing in MoE models. This improvement is reflected not only in the numerical metric but, more importantly, also lays a stronger foundation for expert specialization and collaborative operation within the model.

\subsection{Sensitivity Analysis}

To assess the impact of key hyperparameters in the proposed expert contrastive learning on performance, we adopt a one-factor-at-a-time (OFAT) sensitivity analysis. The backbone is Qwen3 and each MLP layer uses $16$ LoRA experts with top-$4$ sparse routing. The LoRA rank is set to $16$. The expert contrastive learning employs dual projections with an expert memory queue. All other training and inference settings follow Sec .~\ref {sec-exp-setting}. Except for the single hyperparameter under examination, all values are fixed to the defaults. All sensitivity experiments are presented in Fig.~\ref{fig-Sensitivity}, including the projection dimension, queue size, contrastive loss coefficient, and the temperature. As for the sensitivity analysis on the projection dimension $d_h$, it is observed that increasing $d_h$ from $32$ to $64$  yields a clear improvement, while further increasing to $128$ leads to a drop. The results suggest that a too small projection dimension limits the representational capacity for aligning the two views, while a too large dimension increases the capacity of the contrastive heads and the risk of noise fitting, and thus weakens the consistency with the main task. As for the queue size $K$, the negative samples queue exhibits a $U$-shaped performance curve, with $K=8$ performing the best. On the other hand, increasing the queue size to $12$ or $16$ results in performance degradation. However, when the queue is too short ($K=4$), there still exists a performance degeneration. It is probably that $K=4$ is not sufficient enough to keep the diversity of the negative samples. However, when the queue is too long, it will increase the temporal degradation of historical features and cross-batch distribution drift, reducing the discriminative power of InfoNCE. As for the contrastive loss coefficient $\beta$, we still can observe a clear inverted-$U$ shape, in which $\beta=0.01$ achieves the best performance. When $\beta\leq0.001$, the impact from the expert contrastive learning is too weak and thus leads to suboptimal results. Nevertheless, when $\beta\geq0.02$, the conflicts between the contrastive objective and language modeling will be intensified, resulting in the suppression of the main task. Lastly, for the results on the temperature $\tau$, we can observe that there are multiple appropriate choices in a wide range, i.e, between $0.07$ and $0.1$. This aligns with contrastive-learning theory, i.e., the temperature controls the sharpness of the similarity distribution. That is, too small a value makes the distribution overly peaked and suppresses the learning signals from moderately hard negatives, whereas a medium case stabilizes training and enhances discrimination.

\subsection{Case Study}
To gain a deeper understanding of SparseDoctor's advantages in Chinese medical question answering, we selected representative conceptual questions from Chinese medical benchmark test sets and asked two models (SparseDoctor v.s. HuatuoGPT-II) to generate open-ended responses in a natural
dialog style. Two typical cases are presented below to illustrate differences in the understanding of various types of medical knowledge.


\subsubsection{Case 1: Traditional Chinese Medicine (TCM) Pathomechanism} 
In understanding the TCM pathomechanism theory, SparseDoctor accurately identifies "deficiency of righteous qi with lingering pathogenic factors (zheng-xu xie-lian)", as the key pathomechanism that leads to a phenomenon of protracted course of disease. While HuatuoGPT-II incorrectly chooses "stalemate between pathogenic qi and righteous qi (xie-zheng xiang-chi)." Although both involve the relationship between righteous and pathogenic qi, "zheng-xu xie-lian" more accurately reflects the pathological essence of lingering and non-resolving illnesses, demonstrating that the SparseDoctor can precisely comprehend the core concepts of TCM theory.

\subsubsection{Case 2: Drug Identification Knowledge}

For the drug code identification case, the SparseDoctor can correctly identify 654-2 as anisodamine, reflecting accurate mastery of the Chinese drug numbering system. However, HuatuoGPT-II incorrectly identifies it as atropine sulfate. Although both of them are anticholinergic drugs, misidentifying the specific drug could lead to medication errors in clinical practice. Hence, compared to HuatuoGPT-II, SparseDoctor's presents an advantage in the comprehension of drug knowledge.

\section{Conclusion \& Future Works}
In this work, we introduce a new powerful medical LLM, i.e., SparseDoctor, in an architecture-driven way. Compared to the traditional data-driven strategy, such as the HuatuoGPT series, SparseDoctor refers to the efficient and highly sparse LoRA-MoE architecture to expand the model size without the significant increments in computation cost. To further scientifically control the load balancing among different experts, we craft an expert contrastive learning framework to regard the embeddings from the same token as the positive samples and vice versa, which will lead to fine-grained resource allocation among different experts and diversify their functions. Extensive experiments demonstrate the effectiveness of our proposed medical LLM on the medical-related question answering task. In future work, we will investigate the multi-modal chat doctor to comprehend and infer the query from the patient, both from the image data and text data, to pave the way for more precise diagnoses for the medical AI agent.

\bibliographystyle{elsarticle-num}
\bibliography{cas-refs}
\end{document}